\documentclass{llncs}
\usepackage{amsmath}
\usepackage{graphicx}
\usepackage{hyperref}
\usepackage{multirow}
\usepackage{longtable}
\usepackage{array}
\usepackage{amssymb}
\usepackage{amsmath}
\usepackage[ruled,vlined]{algorithm2e}
\def\ie{{\it i.e., }}   % %c'est à dire

\def\R{\mathbb R}  % %l'ensemble des reels= \R
\begin{document}

\title{Image Segmentation Methods for Non-destructive testing Applications}

\author{EL-Hachemi Guerrout \and Ramdane Mahiou \and Randa Boukabene \and Assia Ouali}

\institute{\'Ecole nationale Sup\'erieure en Informatique, Laboratoire LMCS, Oued-Smar, Algiers, Algeria,
	\email{\{e\_guerrout, r\_mahiou, fr\_boukabene, fa\_ouali\}@esi.dz}
}

\maketitle	

\begin{abstract}
	In this paper, we present new image segmentation methods based on hidden Markov random fields (HMRFs) and cuckoo search (CS) variants. HMRFs model the segmentation problem as a minimization of an energy function. CS algorithm is one of the recent powerful optimization techniques. Therefore, five variants of the CS algorithm are used to compute a solution. Through tests, we conduct a study to choose the CS variant with parameters that give good results (execution time and quality of segmentation). CS variants are evaluated and compared with non-destructive testing (NDT) images using a misclassification error (ME) criterion.
\end{abstract}

\begin{keywords}
	Image segmentation; Non-Destructive Testing (NDT) images; Hidden Markov Random Fields; Cuckoo Search Variants; Misclassification Error.
\end{keywords}

%%%%%%%%%%%%%%%%%%%%%%%%%%%%%%%%%%%%%%%%%%%%%%%%%%%%%%%%%%%%%%%%%%%%%%%%%%%%%%%%
\section{INTRODUCTION}
Non Destructive Testing (NDT) is a set of valuable techniques. They have many uses. For example, they are used in manufacturing to inspect materials and extract defective regions without causing damages \cite{cartz1995nondestructive,hellier2001handbook,mix2005introduction,guerrout2014hidden}. NDT methods rely on image analysis where segmentation image is the most important. 

Hidden Markov random fields (HMRFs) \cite{1984-geman-geman,deng2004unsupervised,zhang2001segmentation,guerrout2018hidden} proved their efficiency and robustness in image segmentation. These latter take into consideration the global and local properties of the image. The global properties include the mean and standard deviation values of different pixel classes. The local properties are characterized by the interactions of neighboring pixels. Looking for the segmented image is seen as an optimization problem. Therefore, optimization techniques are used to compute a solution.  

The cuckoo search algorithm is one of the latest powerful nature-inspired meta-heuristic algorithms \cite{yang2009cuckoo}. It is inspired by the behavior of some cuckoo types that put their eggs in the nests of other birds (known as brood parasitism).

In this paper, we shed light on the combinations of HMRFs with five variants of the cuckoo search algorithm that are: standard CS (SCS) \cite{yang2009cuckoo,yang2010engineering,yang2010nature}, improved CS (ICS) \cite{valian2011improved}, auto adaptive modified CS (AACS) \cite{li2015modified}, modified CS (MCS) \cite{walton2011modified} and novel modified (NMCS) \cite{yang2018modified}. The quality of the solution is very sensitive to the choice of parameters. For that reason, we conduct an evaluative study in order to choose parameters that  give a good segmentation. Misclassification error (ME) criterion \cite{sezgin2004survey} measures the difference between the segmentation result and the ground truth (GT). Tests are carried out on NDT (Non-Destructive Testing) image dataset \cite{sezgin2004survey}.

The rest of the paper is organized as follows. In section \ref{hmrf_cs}, we present segmentation methods that are based on HMRF and CS variants. Section \ref{er} is devoted to some experimental results. Finally, we conclude in section \ref{c}.

\section{HMRFs and cuckoo search variants}\label{hmrf_cs}
In image segmentation using HMRFs both image to segment and segmented image are seen as realizations of Markov random variables.

Let $y=({y}_{1},\dots,{y}_{s},{\dots},{y}_{M })$ be the image to segment, $y_s$ is the pixel value of the site $s$ that takes its values in the gray level space $ E_{y}=\{0,\ldots,255\} $ and configurations set $\Omega_y=E_{y}^M$. 

Let $x=({x}_{1},\dots,{x}_{s},{\dots},{x}_{M })$ be the segmented image into $K$ classes, $x_s$ is the class of the site $s$ that takes its values in the discrete space $E_x=\left\{1,{\dots},K\right\}$ and configurations set $\Omega_x=E_{x}^M$.

Let $\mu=(\mu_1,\dots,\mu_j,\dots,\mu_K)$ be the means and $\sigma=(\sigma_1,\dots,\sigma_j,\dots,\sigma_K)$ be the standard deviations of $K$ classes in the segmented image $x=(x_1,\dots,x_s,\dots,x_M)$ \ie
\begin{equation}{\label{sj}}
\begin{array}{l}
\begin{cases}
\mu_j={\frac{1}{|S_j|} \sum_{s \in S_j} y_s}\\\\
\sigma_j=\sqrt{\frac{1}{|S_j|} \sum_{s \in S_j} (y_s-\mu_j)^2}\\\\
S_j=\{s\ |\ x_s=j\}
\end{cases}
\end{array}
\end{equation}

Image segmentation using HMRFs is modeled in \cite{guerrout2018hidden,guerrout2018hiddenpso} as a minimization of an energy function, which is presented in (\ref{e-psi-mu}). We can always compute $x$ through $\mu$ by classifying $y_s$ into the nearest mean $\mu_j$ \ie $x_s=j$ if the nearest mean to $y_s$ is $\mu_j$. Thus, we look for $\mu^*$ instead of $x^*$. The configuration set of $\mu$ is $\Omega_{\mu}=[0\dots255]^K$.
\begin{equation}\label{e-psi-mu}
\begin{array}{l}
\begin{cases}
{\mu}^*=\operatorname*{arg}_{\mu \in \Omega_{\mu}}{\mathit{\min}}\left\{\Psi (\mu)\right\}\\\\
\Psi(\mu) = \sum_{ j = 1 }^{K} \sum\limits_{s \in S_j }{ [\ln (\sigma_j ) + \frac{ (y_s - \mu_j)^2 }{ 2 \sigma_j^2 }] }  +  \frac{B}{T} \sum_{c_2 = \{s,t\}}{ ( 1 - 2 \delta(x_s,x_t) ) } 
\end{cases}
\end{array}
\end{equation}
\noindent where $B$ is a constant, $ T $ is a control parameter called temperature,   
$ \delta $ is Kronecker's delta and $S_j$, $\mu_j$ and $\sigma_j$ are defined in (\ref{sj}).
When  $B>0$, the most likely segmentation corresponds to the constitution of large homogeneous regions. The size of these regions is controlled by the parameter $B$.

To apply unconstrained optimization techniques, we redefine the function $\Psi(\mu)$ for $\mu \in \R^K$ instead of $\mu \in [0 \dots 255]^K$. Therefore, the new function  $\Psi(\mu)$ becomes as follows:  

\begin{equation} \label{psi1}
\begin{array}{l}
\Psi(\mu) =
\begin{cases}
\sum_{ j = 1 }^{K} \sum\limits_{s \in S_j }{ [\ln (\sigma_j ) + \frac{ (y_s - \mu_j)^2 }{ 2 \sigma_j^2 }] } & \multirow{2}{*}{ $\mbox {if} \quad \mu\in [0\dots255]^K$} \\+  \frac{B}{T} \sum_{c_2 = \{s,t\}}{ ( 1 - 2 \delta(x_s,x_t) ) }  & \\\\
+\infty & \mbox {otherwise}
\end{cases}
\end{array}
\end{equation}

%\begin{equation}\label{psi2}
%\Psi(\mu) =\sum_{ j = 1 }^{K} F(\mu_j)\ \ \mbox{where   } \mu_j \in \R 
%\end{equation}
%
%\begin{equation*}
%F(\mu_j)=
%\begin{cases}
%f(0)-u_j*10^3 & \mbox{if } \mu_j < 0\\
%f(\mu_j) & \mbox{if }  \mu_j \in [0\dots255]\\
%f(255)+(u_j-255)*10^3 & \mbox{if } \mu_j > 255
%\end{cases}
%\end{equation*}

%\section{Hidden Markov Random Field and Cuckoo Search (CS) algorithm}
To solve the minimization problem expressed in \ref{psi1}, we have used Cuckoo Search (CS) variants. The principal steps are set out below.

Each egg in a host nest represents a solution $\mu^{i,t}=(\mu^{i,t}_1,\dots,\mu^{i,t}_j,\dots,\mu^{i,t}_K)$ at time $t$.
Let $c^{i,t}=(c^{i,t}_1,,\dots,c^{i,t}_j,\dots,c^{i,t}_K)$ be cuckoo egg generated at time $t$. 
Let $n$ be the number of available host nests (or different solutions). The initial population $\{\mu^{i,0}\}_{i=1,\dots,n}$ is generated by random initialization. 
Let $\mbox{best}^t=(\mbox{best}^t_1,\dots,\mbox{best}^t_j,\dots,\mbox{best}^t_K)$ be the best solution at time $t$. 
\begin{equation}\label{best}
\mbox{best}^t:=\arg\min\limits_{\mu^{i,t}_{i=0,\dots,n} } {\Psi(\mu^{i,t})} 
\end{equation}

Cuckoo Search (CS) algorithm is based on three rules defined in \cite{yang2009cuckoo,yang2010engineering} and proceeds as follows:

\begin{enumerate}
	\item Generating new cuckoos $\{c^{i,t}\}_{i=1,\dots,n}$ can be performed as follows (L\'evy flight):
	\begin{equation}\label{cuckoos}
	c^{i,t}:=\mu^{i,t}+ \alpha \times \mbox{\mbox{step}} \otimes (\mu^{i,t}-\mbox{best}^t)\otimes \mbox{randn}(1,K)
	\end{equation}
	
	\noindent where $\otimes$ means the entry-wise product of two vectors. The $\mbox{randn}(1,K)$ returns $K$ random numbers from a normal distribution with mean 0 and variance 1. 	
	$\alpha>0$ denotes a scaling factor, which is related to the problem under study. In our implementation, the step length ($\mbox{step}$) is calculated by using Mantegna's algorithm (as defined in \cite{yang2010nature}). 
	 
	The nests are updated as follows:
	\begin{equation}\label{update_nests}
	\mu^{i,t}:=
	\begin{cases}
	c^{i,t} & \mbox{if } \Psi(c^{i,t}) \leq \Psi(\mu^{i,t})\\\\
	\mu^{i,t} & \mbox{otherwise} 
	\end{cases}
	\end{equation}
	
	\item 
	A fraction $p_a$ of deficient nests are abandoned.
	New nests $\{v^{i,t}\}_{i=1,\dots,n}$ are generated by biased random walks (crossover operator) as follows:
	
	\begin{equation}\label{new_s_}
	v^{i,t}:=\mu^{i,t}+ \mbox{rand()}\otimes (\mu^{r1,t}-\mu^{r2,t}) \otimes [H(p_a-\mbox{rand()})]
	\end{equation}
	
	\noindent where  $\mu^{r1,t}$ and $\mu^{r2,t}$ are two different solutions selected randomly by random permutation. $H$ is Heaviside function. $\mbox{rand()}$ returns a single uniformly distributed random number in the interval (0,1). $[\  ]$ is a $K$ vector.
	
	\item 
	The best nests will be carried over to the next generations ($t+1$) as follows:
	\begin{equation}\label{best_n}
	\mu^{i,t+1}:=
	\begin{cases}
	v^{i,t} & \mbox{if } \Psi(v^{i,t}) \leq \Psi(\mu^{i,t})\\\\
	\mu^{i,t} & \mbox{otherwise} 
	\end{cases}
	\end{equation}
	
\end{enumerate}

Each variant of CS (SCS, ICS, NMCS, MCS and AACS) manages in a specific way $p_a$, $\alpha$ and $\mbox{step}$.
The standard CS (SCS) uses constant values for the parameters $p_a$ and $\alpha$. As for the improved CS (ICS), the parameters $p_a$ and $\alpha$ are calculated at the time $t$ as expressed in (\ref{Pa_ICS}) and (\ref{alpha_ICS}):

\begin{equation}\label{Pa_ICS}
p_a(t) = {p_a}_{\mbox{max}}-\frac{t}{\mbox{NI}}({p_a}_{\mbox{max}}-{p_a}_{\mbox{min}})
\end{equation}

\begin{equation}\label{alpha_ICS}
\alpha(t) = \alpha_{\mbox{max}} \exp\left(\frac{t\times\ln\left(\frac{\alpha_{\mbox{min}}}{\alpha_{\mbox{max}}}\right)}{\mbox{NI}}\right)
\end{equation}
\noindent where $\mbox{NI}$ is the maximum number of iterations.

Each time, the novel modified CS (NMCS) computes the parameters $p_a$ and $\alpha$ as expressed in (\ref{paa}) and (\ref{alp}).

\begin{equation}\label{paa}
\begin{array}{l}
p_{a}(t)=
\begin{cases}
0.1  &\mbox{if} \qquad p_a(0)/\xi(t)+p_a(0)\times\theta(t) < 0.1\\\\
0.85 & \mbox{if} \qquad p_a(0)/\xi(t)+p_a(0)\times\theta(t) > 0.85 \\\\
p_a(0)/\xi(t)+p_a(0)\times\theta(t) & \mbox{otherwise}  
\end{cases}
\end{array}
\end{equation}

\begin{equation}
\label{alp}
\alpha(t)=\alpha(0)/\xi(t)+\alpha(0)\times\theta(t)
\end{equation}
\noindent where $\xi(t)$ and $\theta(t)$ are the scale conversion factors computed using the speed factor ($\omega_s(t)$) and the aggregation factor ($\omega_a(t)$) as expressed in (\ref{factors}).

\begin{equation}{\label{factors}}
\begin{array}{l}
\begin{cases}
\xi(t)=10^{10\times\tan(\arctan(0.1)\times(2\times\omega_{s}(t)-1))}\\\\
\theta(t)=10^{10\times\tan(\arctan(0.1)\times(2\times\omega_{a}(t)-1))}\\\\
\omega_{s}(t)=\frac{\Psi(\mbox{best}^t)}{\Psi(\mbox{best}^{t-1})}\\\\
\omega_{a}(t)=\frac{\Psi(\mbox{best}^t)}{\frac{1}{n}\sum_{i=1}^{n} (\Psi(\mu^{i,t}))}
\end{cases}
\end{array}
\end{equation}

The Auto adaptive CS (AACS) variant uses a procedure parameter setting that computes for every nest $i$ at the time $t$ a self-adaptive parameter $p_a^i(t)$ (see \cite{li2015modified} for more detail). 
AACS builds a new nests $\{v^{i,t}\}_{i=1,\dots,n}$ using the mutation, crossover and selection operators as follows:

\begin{equation}\label{vi}
\begin{array}{l}
\mbox{If} \quad  \mbox{rand()} >= (1 - t/\mbox{NI})  \qquad \mbox{then}\\
\quad v^{i,t} :=
\begin{cases}
\mbox{best}^t+\varphi\otimes(\mu^{r1,t}-\mu^{r2,t}+\mu^{r3,t}-\mu^{r4,t})   &\mbox{if} \quad  \mbox{rand()} < p^i_a(t) \\\\
\mu^{i,t} & \mbox{otherwise}  
\end{cases}
\\
\mbox{Else} \\
\quad v^{i,t} :=
\begin{cases}
\mu^{r1,t}+\varphi\otimes(\mu^{r2,t}-\mu^{r3,t})   &\mbox{if} \quad \mbox{rand()} < p^i_a(t) \\\\
\mu^{i,t} & \mbox{otherwise}  
\end{cases}\\
\mbox{End If} 
\end{array}
\end{equation}
\noindent where  $\mu^{r1,t}$, $\mu^{r2,t}$, $\mu^{r3,t}$ and $\mu^{r4,t}$ are four different solutions selected randomly by random permutation. $\mbox{rand()}$ returns a single uniformly distributed random number in the interval (0,1). $\varphi$ is the scale factor that represents a Gaussian distribution with mean $0.5$ and standard deviation $0.1$.

In Algorithm \ref{3_algo}, we summarize four combinations: HMRF-SCS, HMRF-ICS, HMRF-NMCS, and HMRF-AACS. 

\begin{algorithm}[h] \label{3_algo}
	\SetAlgoLined
	The objective function is $\Psi(\mu), \mu=(\mu_{1},\dots,\mu_{K})\in \R^K$
	
	Generate the initial population of $n$ host nests $\{\mu^{i,0}\}_{i=1,\dots,n}$
	
	\While{($t < \mbox{NI} $) }{
		Compute the best solution $\mbox{best}^t$ using  (\ref{best})
		
		\If{ICS}{
			compute $\alpha(t)$ using (\ref{alpha_ICS})
		}
		\If{NMCS}{
		compute $\alpha(t)$ using (\ref{alp})
	    }
	    Generating new cuckoos $\{c^{i,t}\}_{i=1,\dots,n}$
		using  (\ref{cuckoos})
		
		Update nests $\{\mu^{i,t}\}_{i=1,\dots,n}$ using  (\ref{update_nests}) 
		
		\If{ICS}{
			compute $p_a(t)$ using (\ref{Pa_ICS})
		}
	    \If{NMCS}{
	    compute $p_a(t)$ using (\ref{paa})
	    }
        \If{AACS}{
        	compute $p^i_a(t)$ using the procedure parameter setting
        }
		A fraction $p_a$ of deficient nests are abandoned
		
		\eIf{AACS}{
			New nests $\{v^{i,t}\}_{i=1,\dots,n}$ are built using (\ref{vi})
		}
	    {
	    	New nests $\{v^{i,t}\}_{i=1,\dots,n}$ are built using (\ref{new_s_})
	    }
		
		Keep best nests to the next generations ($t+1$) using (\ref{best_n})
		
		$t:=t+1$
	}
	\caption{HMRF-SCS, HMRF-ICS, HMRF-NMCS, and HMRF-AACS algorithms.}
\end{algorithm}

The modified CS (MCS) divides nests into two groups. The first group is the top nests with high fitness ($25\%$ of nests). The second group is the abandoned nest with low fitness ($75\%$ of nests). 
The HMRF-MCS combination is summarized in Algorithm \ref{MCS}.\\

\begin{algorithm}[t]\label{MCS}
	\SetAlgoLined
 The objective function is $\Psi(\mu), \mu=(\mu_{1},\dots,\mu_{K})\in \R^K$

Generate the initial population of $n$ host nests $\{\mu^{i,0}\}_{i=1,\dots,n}$

\While{($t < \mbox{NI} $) }{
		Split eggs into $N1=$top nests (25\%) and 
		$N2=$abandoned nests (75\%)
		
		\ForEach{i $\in$ N2}{
			Get new cuckoos using  \ref{cuckoos}~ with~ $\alpha=\frac{1}{\sqrt{t}}$\\
			Replace the nest $i$ by the new one. 
			
		}
		
		\ForEach{i $\in$ N1}{

			pick a nest $j \in N1$ randomly
			
			\eIf{$\mu^{i,t} = \mu^{j,t}$}{
				Get new cuckoos using  \ref{cuckoos}~ with~ $\alpha=\frac{1}{t^2}$\\
				Choose a random nest $l$ from all nests\\
				Update the nest $l$ by the new one using  \ref{update_nests} 
				
			}{
				dx $= | \mu^{i,t} - \mu^{j,t} |/\phi $ with $\phi=(1+\sqrt{5})/2$\\
				Move distance dx from the worst nest to the best nest to generate a new one\\
				Choose a random nest $l$ from all nests\\
				Update the nest $l$ by the new one using 
			    \ref{update_nests} 
				
			}
		}
		
		Update $t:=t+1$
	}
	\caption{HMRF-MCS algorithm.}
\end{algorithm}	

\vspace{2.0mm}

\section{Experimental Results}\label{er}
To have a meaningful study and comparison,
we have performed our tests in two steps. In the first step, we have conducted an evaluative study for each CS variant to select parameters that give a good segmentation. Each algorithm is tested upon 25 images of the NDT dataset \cite{sezgin2004survey}. The values of parameter $n$ (number of nests) tested are $5, 10, 15, 20, 25, \mbox{ and } 30$. The values of parameter $NI$ (number of iterations) tested are $50 \mbox{ and } 100$. The values of parameter $T$ (temperature) tested are $2 \mbox{ and } 3$. The other parameters ($\alpha, p_a, \phi, \varphi, \mbox{ and } \mbox{ step}$)  are used as recommended in the basic articles \cite{yang2009cuckoo,valian2011improved,li2015modified,walton2011modified,yang2018modified}.  
In the second step,
we carried out a comparative study of different combinations with the best parameters selected in the first step. The comparison focuses on the execution time and misclassification error results. 

\subsection{Misclassification error criterion (ME)}
In the case of binary images, ME measures the difference between segmented image and ground truth (GT). The binary images are constituted by a foreground and a background. ME provides the percentage of misclassification pixels as follows\cite{sezgin2004survey}:
\begin{equation}
ME=1-\frac{|B_O \bigcap B_T| + |F_O \bigcap F_T|}{|B_O| + |F_O|}
\end{equation}

\noindent where $F_O$ and $B_O$ indicate foreground and background of ground truth (GT), $F_T$ and $B_T$ indicate foreground and background of the segmented image. ME equals zero that means the perfect match between segmented image and its ground-truth.

\subsection{Tests context}\label{tc}
Algorithms are implemented in MATLAB 2017a on a computer with Intel Core i7 1.8 GHz CPU, 8G RAM, and Microsoft Windows 10. 
%Table \ref{p} shows the different parameters used in our tests.

%\begin{table}[t]
%	\centering 
%	\begin{tabular}{|l|c|}
%		\hline
%		Parameters &  values\\
%		\hline
%		The number of nests & $n\in\{5,10,15,20,25,30\}$\\
%		\hline
%		Temperature & $T\in\{2,3\}$\\
%		\hline
%		The number of iterations & $NI\in\{50,100\}$\\
%		\hline
%	\end{tabular}
%	\caption{Parameters used in the evaluative study.}
%	\label{p}
%\end{table}

\subsection{Results}
After the evaluative study, we have selected for each algorithm the parameters that give the best results, which are shown in Table \ref{me1}. Mean ME values and mean execution time are calculated using all NDT images.
In Table \ref{me1}, the best results are given in bold type. 

\begin{table}[h]
	\centering
	\begin{tabular}{|l|l|l|l|}
	   \hline
		Methods & Parameters & ME & Time(s)  \\
		\hline
		SCS & n=20, NI=100, T=2 & 0.067 & 13.634  \\
		\hline
		ICS & n=20, NI=100, T=2 & \textbf{0.046} & 13.516  \\
		\hline
		AACS & n=25, NI=100, T=2 & 0.055 & 26.313  \\
		\hline
		MCS & n=5, NI=100, T=3 & 0.058 & \textbf{1.329}  \\
		\hline
		NMCS & n=5, NI=50, T=2 & 0.066 & 1.633 \\
		\hline		
	\end{tabular}
	\caption{Parameters, mean ME values, and mean execution time.}
	\label{me1}
\end{table}

\begin{figure}[t]
    \centering
    \newlength{\iw}
    \newlength{\ih}	
    \renewcommand{\angle}{0}
    \setlength{\iw}{12cm}
    \setlength{\ih}{4.96cm}
	\includegraphics[width=\iw,height=\ih,angle=\angle]{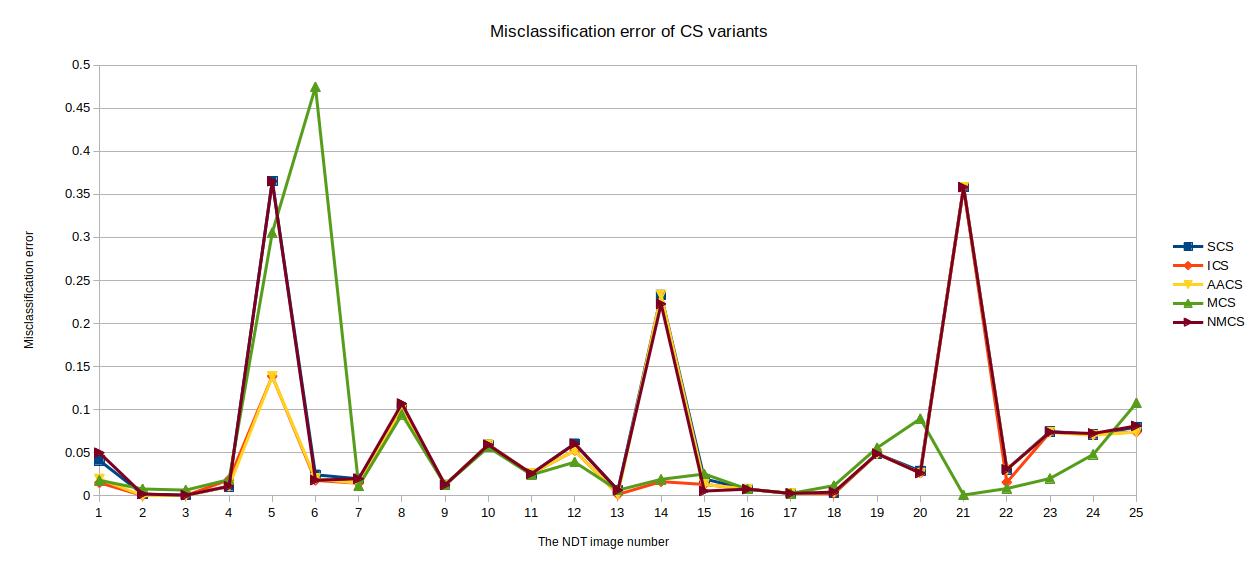}
	\caption{Misclassification error of each algorithm compared to images.}
	\label{mef}
\end{figure}

\begin{figure}[t]
    \centering
    \renewcommand{\angle}{0}
    \setlength{\iw}{12cm}
    \setlength{\ih}{5cm}
	\includegraphics[width=\iw,height=\ih,angle=\angle]{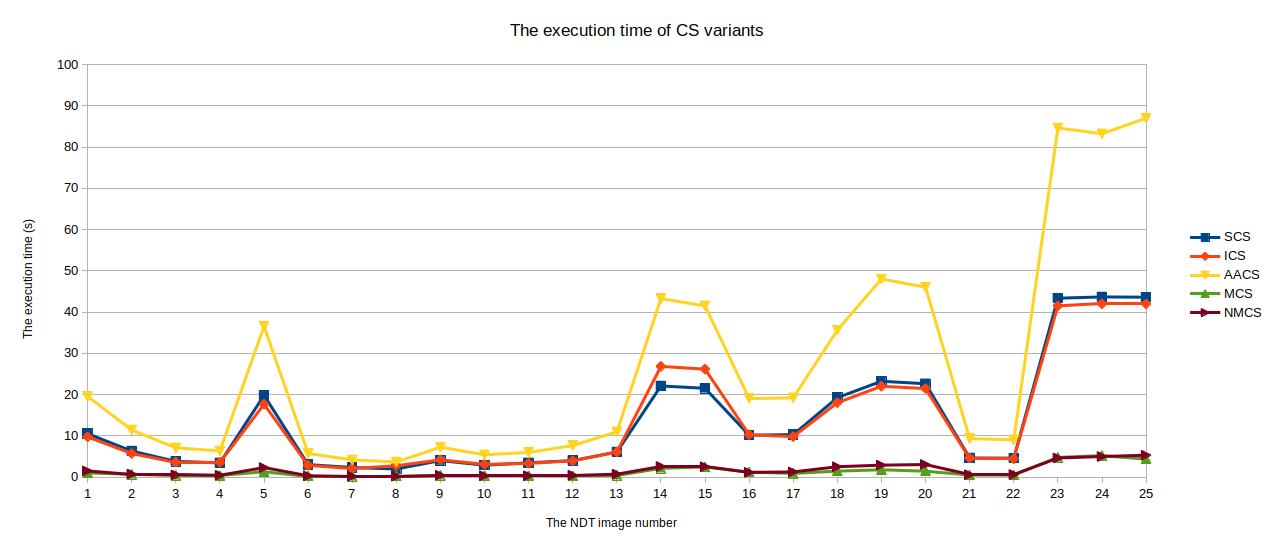}
	\caption{The execution time of each algorithm with the best parameters compared to images.}
	\label{tf}
\end{figure}

\begin{figure}[t]
    \renewcommand{\arraystretch}{0.5}
\begin{center}

	\renewcommand{\angle}{0}
	\setlength{\iw}{1.5cm}
	\setlength{\ih}{0.7cm}
	
	\begin{tabular}{|c|c|c|c|c|c|c|c|}
		\hline
		No & Image & GT & SCS & ICS & AACS & MCS & NMCS \\
		
	\hline
	1 & \includegraphics[width=\iw,height=\ih,angle=\angle]{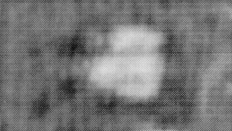}&
	\includegraphics[width=\iw,height=\ih,angle=\angle]{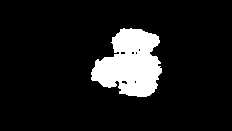} &
	\includegraphics[width=\iw,height=\ih,angle=\angle]{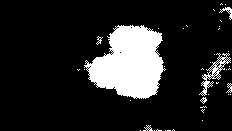}&
	\includegraphics[width=\iw,height=\ih,angle=\angle]{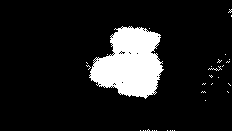}&
	\includegraphics[width=\iw,height=\ih,angle=\angle]{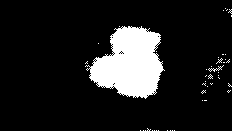}&
	\includegraphics[width=\iw,height=\ih,angle=\angle]{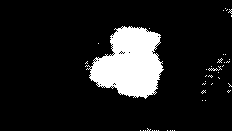}&
	\includegraphics[width=\iw,height=\ih,angle=\angle]{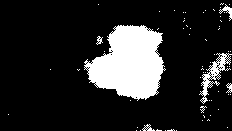} \\

	\hline
	2&\includegraphics[width=\iw,height=\ih,angle=\angle]{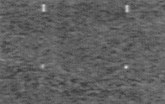}&
	\includegraphics[width=\iw,height=\ih,angle=\angle]{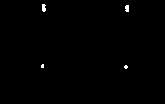} &
	\includegraphics[width=\iw,height=\ih,angle=\angle]{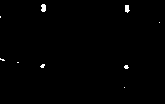}&
	\includegraphics[width=\iw,height=\ih,angle=\angle]{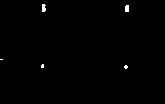}&
	\includegraphics[width=\iw,height=\ih,angle=\angle]{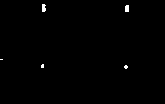}&
	\includegraphics[width=\iw,height=\ih,angle=\angle]{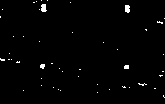}&
	\includegraphics[width=\iw,height=\ih,angle=\angle]{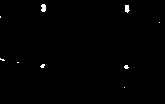}\\

	\hline
	3&\includegraphics[width=\iw,height=\ih,angle=\angle]{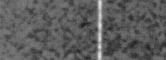}&
	\includegraphics[width=\iw,height=\ih,angle=\angle]{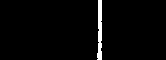} &
	\includegraphics[width=\iw,height=\ih,angle=\angle]{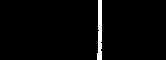}&
	\includegraphics[width=\iw,height=\ih,angle=\angle]{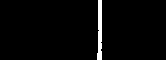}&
	\includegraphics[width=\iw,height=\ih,angle=\angle]{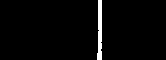}&
	\includegraphics[width=\iw,height=\ih,angle=\angle]{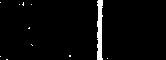}&
	\includegraphics[width=\iw,height=\ih,angle=\angle]{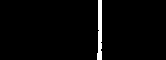}\\

	\hline
	4&\includegraphics[width=\iw,height=\ih,angle=\angle]{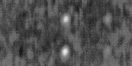}&
	\includegraphics[width=\iw,height=\ih,angle=\angle]{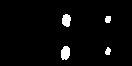} &
	\includegraphics[width=\iw,height=\ih,angle=\angle]{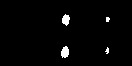}&
	\includegraphics[width=\iw,height=\ih,angle=\angle]{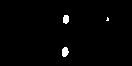}&
	\includegraphics[width=\iw,height=\ih,angle=\angle]{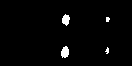}&
	\includegraphics[width=\iw,height=\ih,angle=\angle]{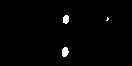}&
	\includegraphics[width=\iw,height=\ih,angle=\angle]{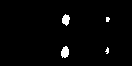}\\

	\hline
	5&\includegraphics[width=\iw,height=\ih,angle=\angle]{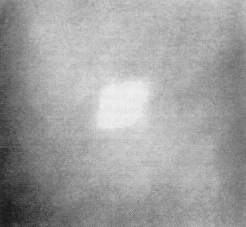}&
	\includegraphics[width=\iw,height=\ih,angle=\angle]{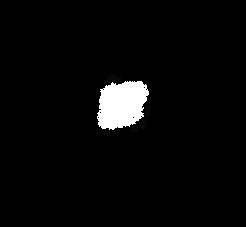} &
	\includegraphics[width=\iw,height=\ih,angle=\angle]{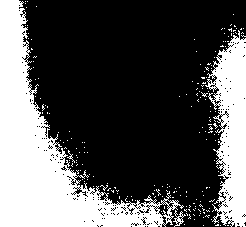}&
	\includegraphics[width=\iw,height=\ih,angle=\angle]{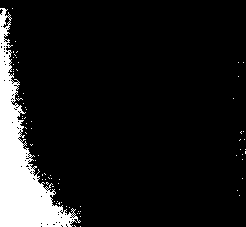}&
	\includegraphics[width=\iw,height=\ih,angle=\angle]{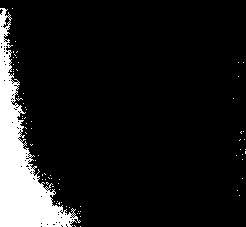}&
	\includegraphics[width=\iw,height=\ih,angle=\angle]{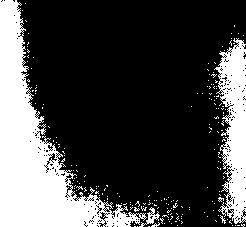}&
	\includegraphics[width=\iw,height=\ih,angle=\angle]{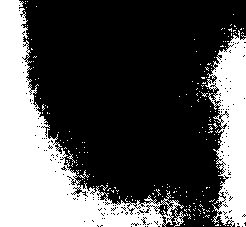}\\

	\hline
	6&\includegraphics[width=\iw,height=\ih,angle=\angle]{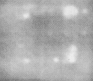}&
	\includegraphics[width=\iw,height=\ih,angle=\angle]{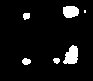} &
	\includegraphics[width=\iw,height=\ih,angle=\angle]{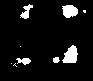}&
	\includegraphics[width=\iw,height=\ih,angle=\angle]{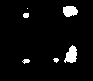}&
	\includegraphics[width=\iw,height=\ih,angle=\angle]{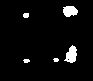}&
	\includegraphics[width=\iw,height=\ih,angle=\angle]{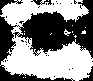}&
	\includegraphics[width=\iw,height=\ih,angle=\angle]{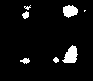}\\

	\hline
	7&\includegraphics[width=\iw,height=\ih,angle=\angle]{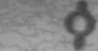}&
	\includegraphics[width=\iw,height=\ih,angle=\angle]{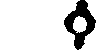} &
	\includegraphics[width=\iw,height=\ih,angle=\angle]{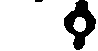}&
	\includegraphics[width=\iw,height=\ih,angle=\angle]{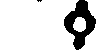}&
	\includegraphics[width=\iw,height=\ih,angle=\angle]{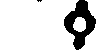}&
	\includegraphics[width=\iw,height=\ih,angle=\angle]{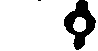}&
	\includegraphics[width=\iw,height=\ih,angle=\angle]{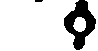}\\

	\hline
	8&\includegraphics[width=\iw,height=\ih,angle=\angle]{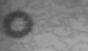}&
	\includegraphics[width=\iw,height=\ih,angle=\angle]{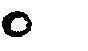} &
	\includegraphics[width=\iw,height=\ih,angle=\angle]{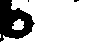}&
	\includegraphics[width=\iw,height=\ih,angle=\angle]{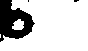}&
	\includegraphics[width=\iw,height=\ih,angle=\angle]{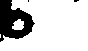}&
	\includegraphics[width=\iw,height=\ih,angle=\angle]{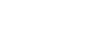}&
	\includegraphics[width=\iw,height=\ih,angle=\angle]{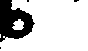}\\

	\hline
	9&\includegraphics[width=\iw,height=\ih,angle=\angle]{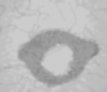}&
	\includegraphics[width=\iw,height=\ih,angle=\angle]{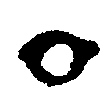} &
	\includegraphics[width=\iw,height=\ih,angle=\angle]{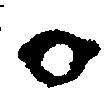}&
	\includegraphics[width=\iw,height=\ih,angle=\angle]{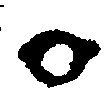}&
	\includegraphics[width=\iw,height=\ih,angle=\angle]{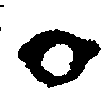}&
	\includegraphics[width=\iw,height=\ih,angle=\angle]{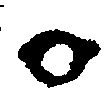}&
	\includegraphics[width=\iw,height=\ih,angle=\angle]{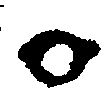}\\

	\hline
	10&\includegraphics[width=\iw,height=\ih,angle=\angle]{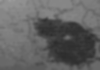}&
	\includegraphics[width=\iw,height=\ih,angle=\angle]{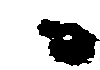} &
	\includegraphics[width=\iw,height=\ih,angle=\angle]{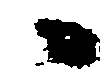}&
	\includegraphics[width=\iw,height=\ih,angle=\angle]{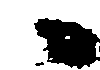}&
	\includegraphics[width=\iw,height=\ih,angle=\angle]{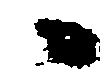}&
	\includegraphics[width=\iw,height=\ih,angle=\angle]{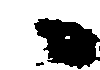}&
	\includegraphics[width=\iw,height=\ih,angle=\angle]{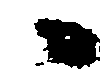}\\

		\hline
	11&\includegraphics[width=\iw,height=\ih,angle=\angle]{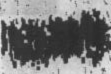}&
	\includegraphics[width=\iw,height=\ih,angle=\angle]{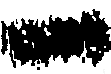} &
	\includegraphics[width=\iw,height=\ih,angle=\angle]{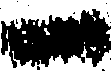}&
	\includegraphics[width=\iw,height=\ih,angle=\angle]{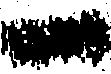}&
	\includegraphics[width=\iw,height=\ih,angle=\angle]{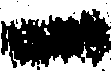}&
	\includegraphics[width=\iw,height=\ih,angle=\angle]{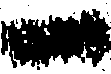}&
	\includegraphics[width=\iw,height=\ih,angle=\angle]{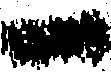}\\

			\hline
	12&\includegraphics[width=\iw,height=\ih,angle=\angle]{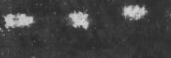}&
	\includegraphics[width=\iw,height=\ih,angle=\angle]{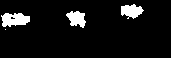} &
	\includegraphics[width=\iw,height=\ih,angle=\angle]{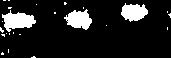}&
	\includegraphics[width=\iw,height=\ih,angle=\angle]{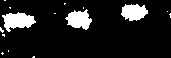}&
	\includegraphics[width=\iw,height=\ih,angle=\angle]{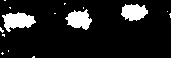}&
	\includegraphics[width=\iw,height=\ih,angle=\angle]{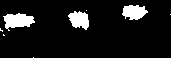}&
	\includegraphics[width=\iw,height=\ih,angle=\angle]{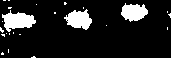}\\

	\hline
	13&\includegraphics[width=\iw,height=\ih,angle=\angle]{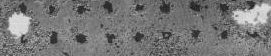}&
	\includegraphics[width=\iw,height=\ih,angle=\angle]{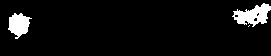} &
	\includegraphics[width=\iw,height=\ih,angle=\angle]{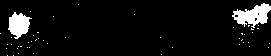}&
	\includegraphics[width=\iw,height=\ih,angle=\angle]{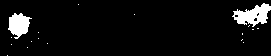}&
	\includegraphics[width=\iw,height=\ih,angle=\angle]{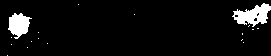}&
	\includegraphics[width=\iw,height=\ih,angle=\angle]{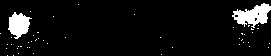}&
	\includegraphics[width=\iw,height=\ih,angle=\angle]{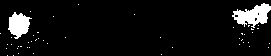}\\
	
	\hline
	14&\includegraphics[width=\iw,height=\ih,angle=\angle]{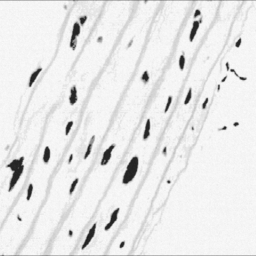}&
	\includegraphics[width=\iw,height=\ih,angle=\angle]{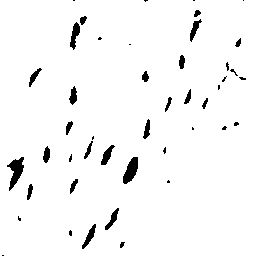} &
	\includegraphics[width=\iw,height=\ih,angle=\angle]{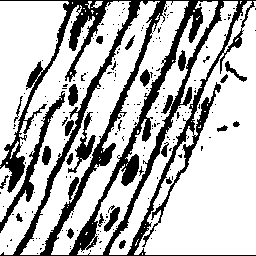}&
	\includegraphics[width=\iw,height=\ih,angle=\angle]{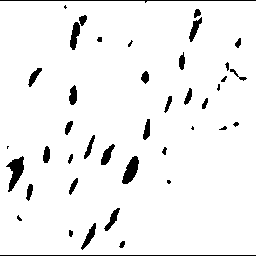}&
	\includegraphics[width=\iw,height=\ih,angle=\angle]{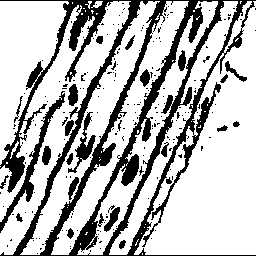}&
	\includegraphics[width=\iw,height=\ih,angle=\angle]{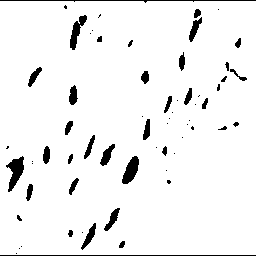}&
	\includegraphics[width=\iw,height=\ih,angle=\angle]{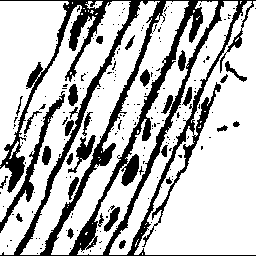}\\

	\hline
15&\includegraphics[width=\iw,height=\ih,angle=\angle]{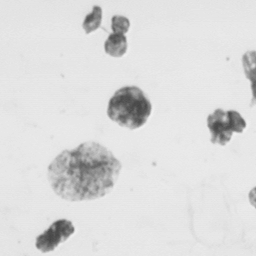}&
\includegraphics[width=\iw,height=\ih,angle=\angle]{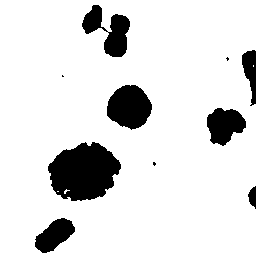} &
\includegraphics[width=\iw,height=\ih,angle=\angle]{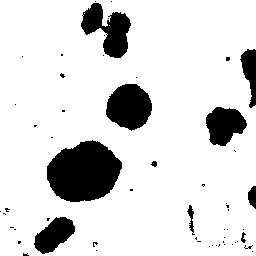}&
\includegraphics[width=\iw,height=\ih,angle=\angle]{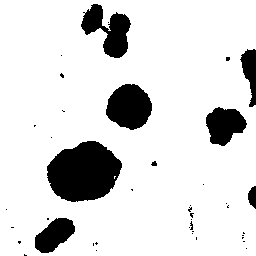}&
\includegraphics[width=\iw,height=\ih,angle=\angle]{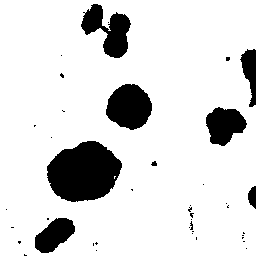}&
\includegraphics[width=\iw,height=\ih,angle=\angle]{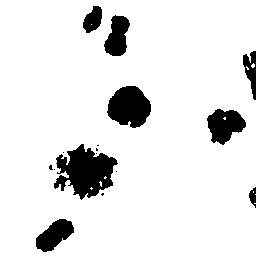}&
\includegraphics[width=\iw,height=\ih,angle=\angle]{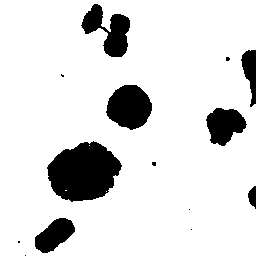}\\

	\hline
16&\includegraphics[width=\iw,height=\ih,angle=\angle]{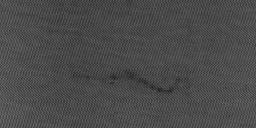}&
\includegraphics[width=\iw,height=\ih,angle=\angle]{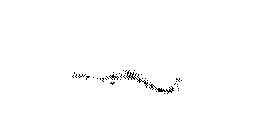} &
\includegraphics[width=\iw,height=\ih,angle=\angle]{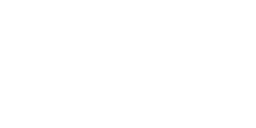}&
\includegraphics[width=\iw,height=\ih,angle=\angle]{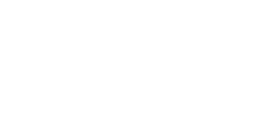}&
\includegraphics[width=\iw,height=\ih,angle=\angle]{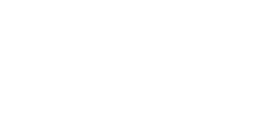}&
\includegraphics[width=\iw,height=\ih,angle=\angle]{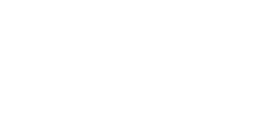}&
\includegraphics[width=\iw,height=\ih,angle=\angle]{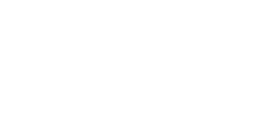}\\

	\hline
17&\includegraphics[width=\iw,height=\ih,angle=\angle]{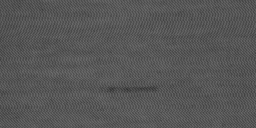}&
\includegraphics[width=\iw,height=\ih,angle=\angle]{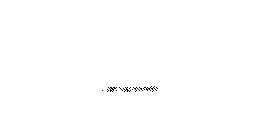} &
\includegraphics[width=\iw,height=\ih,angle=\angle]{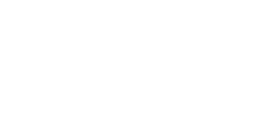}&
\includegraphics[width=\iw,height=\ih,angle=\angle]{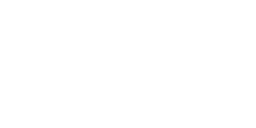}&
\includegraphics[width=\iw,height=\ih,angle=\angle]{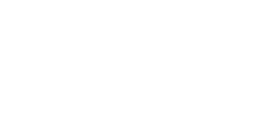}&
\includegraphics[width=\iw,height=\ih,angle=\angle]{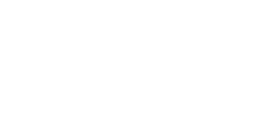}&
\includegraphics[width=\iw,height=\ih,angle=\angle]{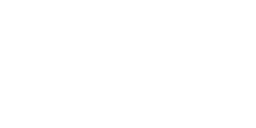}\\	

	\hline
18&\includegraphics[width=\iw,height=\ih,angle=\angle]{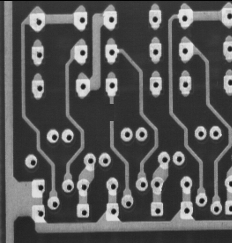}&
\includegraphics[width=\iw,height=\ih,angle=\angle]{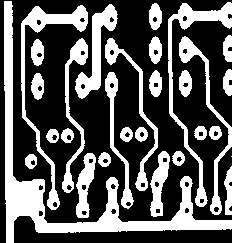} &
\includegraphics[width=\iw,height=\ih,angle=\angle]{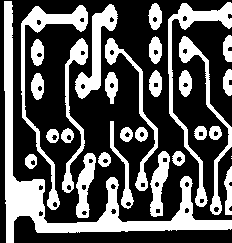}&
\includegraphics[width=\iw,height=\ih,angle=\angle]{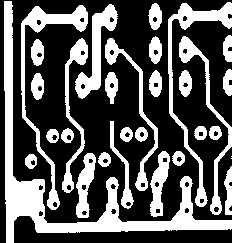}&
\includegraphics[width=\iw,height=\ih,angle=\angle]{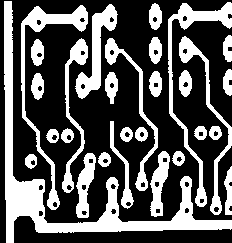}&
\includegraphics[width=\iw,height=\ih,angle=\angle]{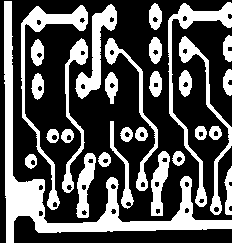}&
\includegraphics[width=\iw,height=\ih,angle=\angle]{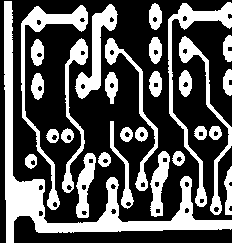}\\	

	\hline
19&\includegraphics[width=\iw,height=\ih,angle=\angle]{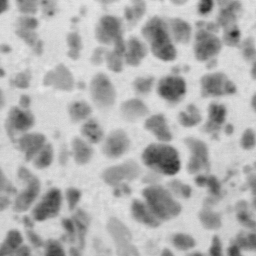}&
\includegraphics[width=\iw,height=\ih,angle=\angle]{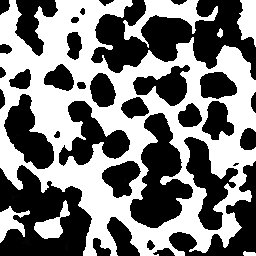} &
\includegraphics[width=\iw,height=\ih,angle=\angle]{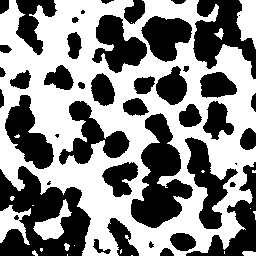}&
\includegraphics[width=\iw,height=\ih,angle=\angle]{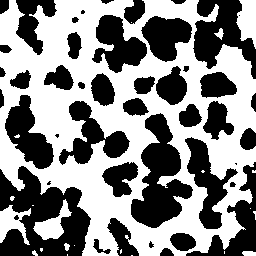}&
\includegraphics[width=\iw,height=\ih,angle=\angle]{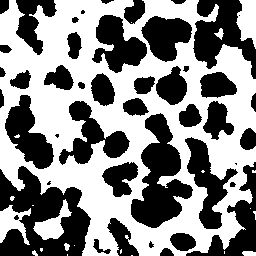}&
\includegraphics[width=\iw,height=\ih,angle=\angle]{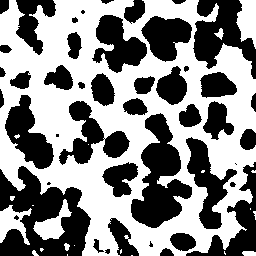}&
\includegraphics[width=\iw,height=\ih,angle=\angle]{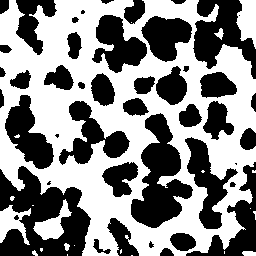}\\	
	
	\hline
	20&\includegraphics[width=\iw,height=\ih,angle=\angle]{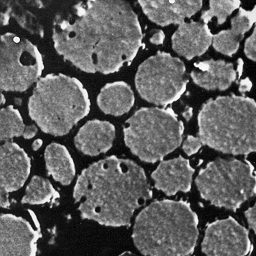}&
	\includegraphics[width=\iw,height=\ih,angle=\angle]{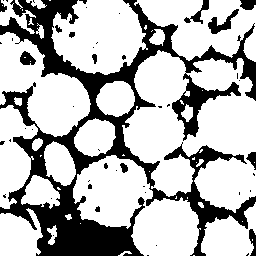} &
	\includegraphics[width=\iw,height=\ih,angle=\angle]{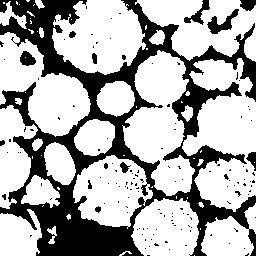}&
	\includegraphics[width=\iw,height=\ih,angle=\angle]{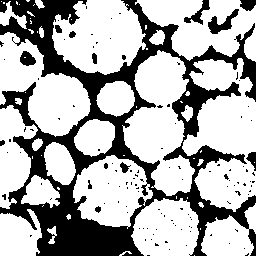}&
	\includegraphics[width=\iw,height=\ih,angle=\angle]{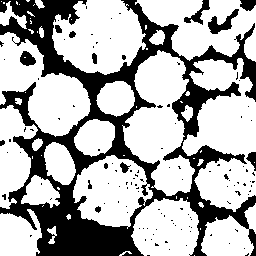}&
	\includegraphics[width=\iw,height=\ih,angle=\angle]{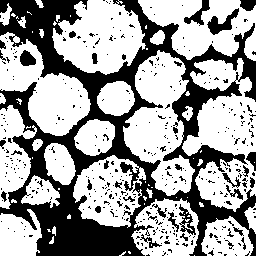}&
	\includegraphics[width=\iw,height=\ih,angle=\angle]{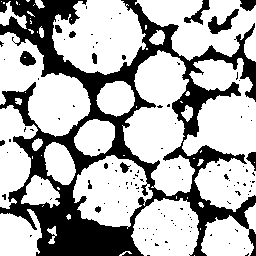}\\

	\hline

	21&\includegraphics[width=\iw,height=\ih,angle=\angle]{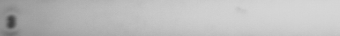}&
	\includegraphics[width=\iw,height=\ih,angle=\angle]{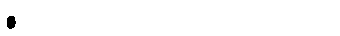} &
	\includegraphics[width=\iw,height=\ih,angle=\angle]{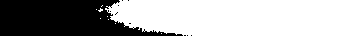}&
	\includegraphics[width=\iw,height=\ih,angle=\angle]{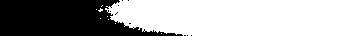}&
	\includegraphics[width=\iw,height=\ih,angle=\angle]{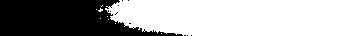}&
	\includegraphics[width=\iw,height=\ih,angle=\angle]{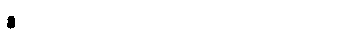}&
	\includegraphics[width=\iw,height=\ih,angle=\angle]{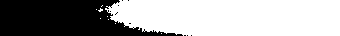}\\

	\hline

	22&\includegraphics[width=\iw,height=\ih,angle=\angle]{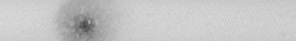}&
\includegraphics[width=\iw,height=\ih,angle=\angle]{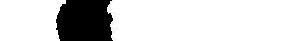} &
\includegraphics[width=\iw,height=\ih,angle=\angle]{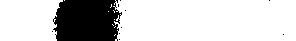}&
\includegraphics[width=\iw,height=\ih,angle=\angle]{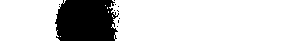}&
\includegraphics[width=\iw,height=\ih,angle=\angle]{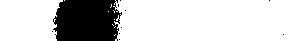}&
\includegraphics[width=\iw,height=\ih,angle=\angle]{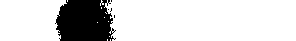}&
\includegraphics[width=\iw,height=\ih,angle=\angle]{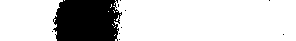}\\

\hline	
	
	23&\includegraphics[width=\iw,height=\ih,angle=\angle]{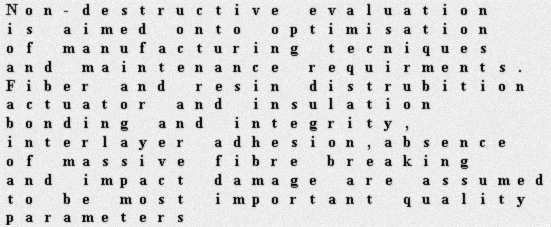}&
	\includegraphics[width=\iw,height=\ih,angle=\angle]{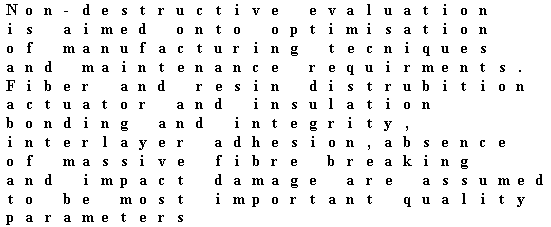} &
	\includegraphics[width=\iw,height=\ih,angle=\angle]{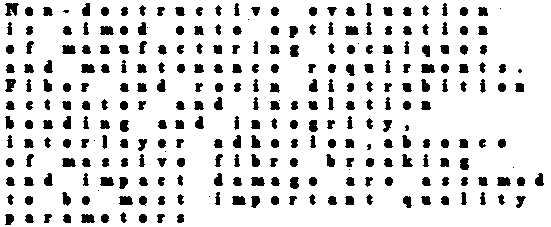}&
	\includegraphics[width=\iw,height=\ih,angle=\angle]{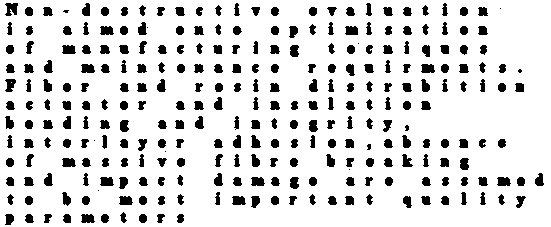}&
	\includegraphics[width=\iw,height=\ih,angle=\angle]{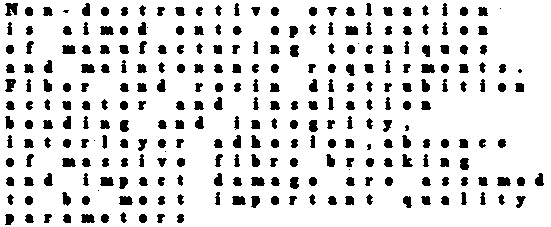}&
	\includegraphics[width=\iw,height=\ih,angle=\angle]{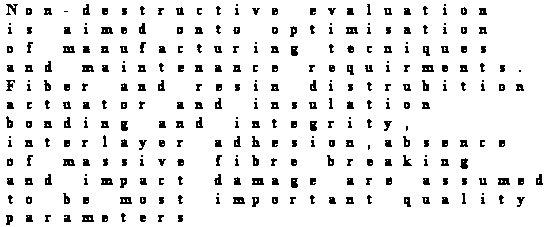}&
	\includegraphics[width=\iw,height=\ih,angle=\angle]{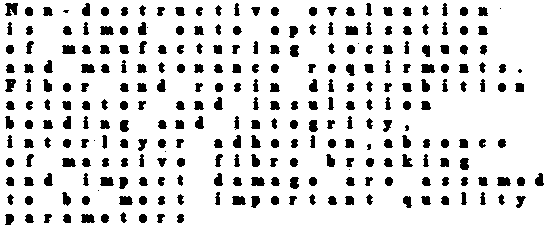}\\
	
	\hline	
	
	24&\includegraphics[width=\iw,height=\ih,angle=\angle]{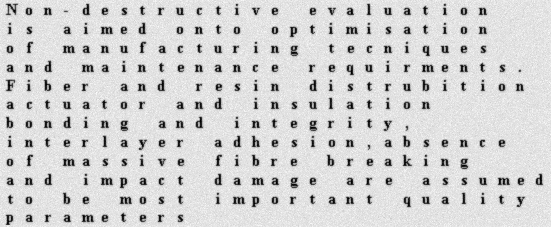}&
	\includegraphics[width=\iw,height=\ih,angle=\angle]{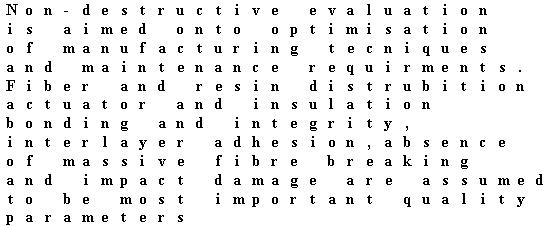} &
	\includegraphics[width=\iw,height=\ih,angle=\angle]{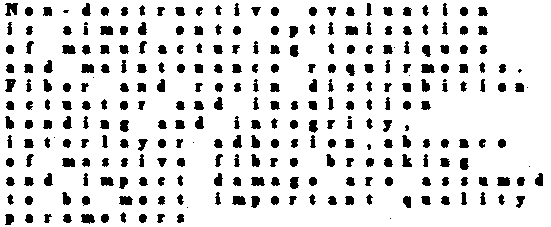}&
	\includegraphics[width=\iw,height=\ih,angle=\angle]{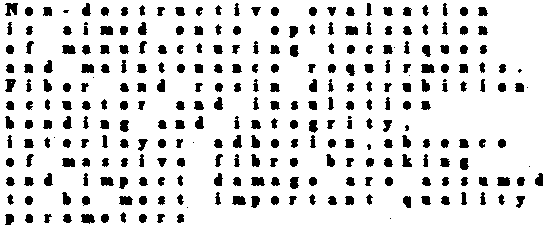}&
	\includegraphics[width=\iw,height=\ih,angle=\angle]{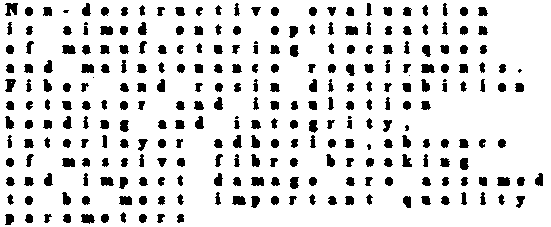}&
	\includegraphics[width=\iw,height=\ih,angle=\angle]{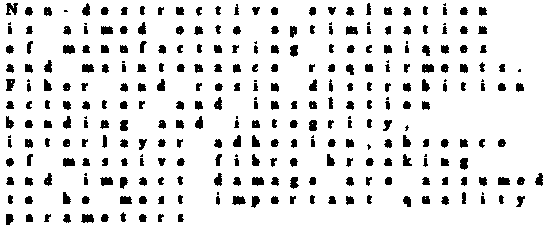}&
	\includegraphics[width=\iw,height=\ih,angle=\angle]{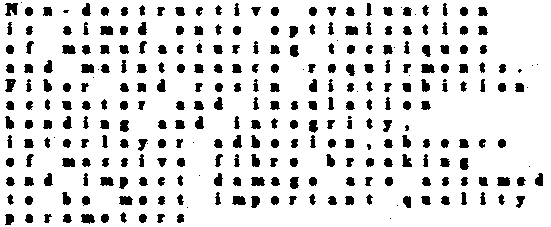}\\
	
	\hline	
	
	25&\includegraphics[width=\iw,height=\ih,angle=\angle]{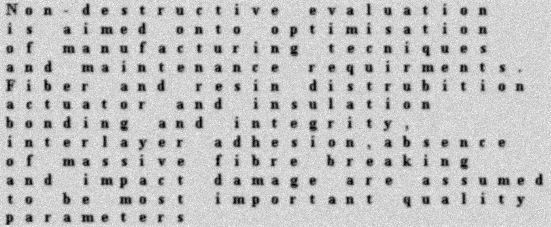}&
	\includegraphics[width=\iw,height=\ih,angle=\angle]{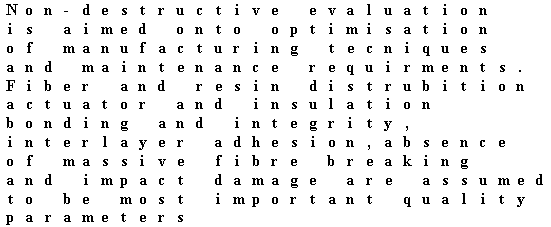} &
	\includegraphics[width=\iw,height=\ih,angle=\angle]{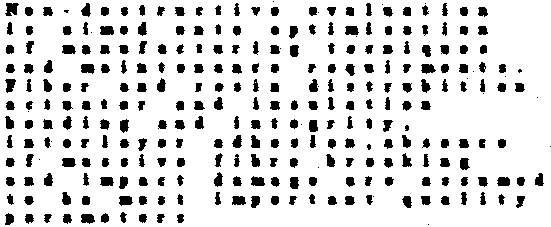}&
	\includegraphics[width=\iw,height=\ih,angle=\angle]{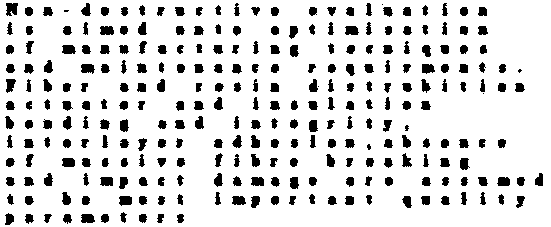}&
	\includegraphics[width=\iw,height=\ih,angle=\angle]{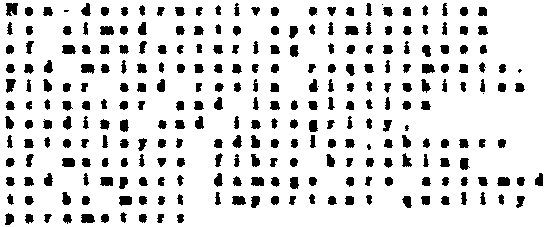}&
	\includegraphics[width=\iw,height=\ih,angle=\angle]{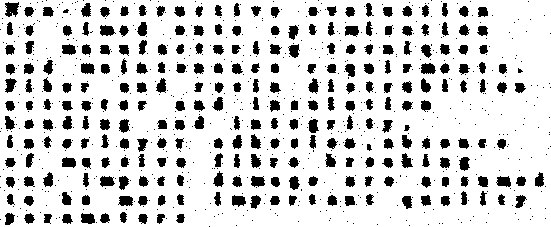}&
	\includegraphics[width=\iw,height=\ih,angle=\angle]{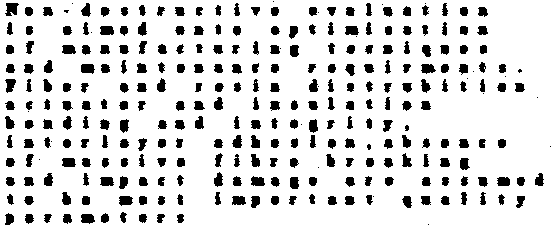}\\

	\hline
				
		\end{tabular}
		\caption{Visual results of each algorithm.}
		\label{vr}
\end{center}
	\end{figure}

Figures \ref{mef} and \ref{tf} provide a comparison between algorithms on ME and execution time respectively compared to NDT images. 
The algorithms are executed with the parameters presented in Table \ref{me1}.
Figure \ref{vr} gives the visual results, which provide an integrated view with ME.    

\section{Conclusion and discussion}\label{c}
In this paper, we have presented new approaches that combine hidden Markov random fields and cuckoo search variants to perform segmentation. 
Then, we have conducted an evaluative study for each CS variant in order to select good parameters. After that,
we have carried out a comparative study between different combinations. Each algorithm is tested upon 25 images where ground truth (GT) segmentation is known. The comparison focuses on the execution time and misclassification error (ME). 
HMRC-ICS combination shows the best results $ME=0.046$ and 21 segmented images visually close to the ground truth. On the other hand, HMRF-MCS presents an interesting execution time and ME not far from the best results. 
To make a fair choice, prospectively we will invest in a statistical study of the parameters: $\mbox{step}$, $\alpha$, $\phi$, $\varphi$,  and $p_a$.
\bibliographystyle{splncs03}
\bibliography{strings}
\end{document}